\pdfoutput=1

\documentclass[11pt]{article}

\usepackage{EMNLP2022}

\usepackage{times}
\usepackage{latexsym}

\usepackage[T1]{fontenc}

\usepackage[utf8]{inputenc}

\usepackage{microtype}

\usepackage{inconsolata}
\usepackage{algorithm}
\usepackage{algpseudocode}
\usepackage{amsmath}
\usepackage{amssymb}
\usepackage{amsmath}
\usepackage{arydshln}
\usepackage{ascii}
\usepackage{bm}
\usepackage{booktabs}
\usepackage{colortbl}
\usepackage{color}
\usepackage{CJK}
\usepackage{dsfont}
\usepackage{enumitem}
\usepackage{forest}
\usepackage{graphics}
\usepackage{graphicx}
\usepackage{latexsym}
\usepackage{multirow}
\usepackage{mwe}
\usepackage{makecell}
\usepackage{microtype}
\usepackage{subfig}
\usepackage{times}
\usepackage{tikz}
\usepackage{tabularx}
\usepackage{url}
\usepackage{wrapfig}

\definecolor{sred}{RGB}{203, 64, 46}
\definecolor{sblue}{RGB}{44, 73, 135}
\definecolor{sgreen}{RGB}{37, 100, 28}

\title{A Template-based Method for Constrained Neural Machine Translation}

\author{
    Shuo Wang$^1$ \ 
    Peng Li$^{2*}$ \ 
    Zhixing Tan$^{6}$ \ 
    Zhaopeng Tu$^7$ \ 
    Maosong Sun$^{1,4}$ \ 
    Yang Liu$^{1,2,3,4,5}$\thanks{\ \ Corresponding authors: P.Li (\href{mailto:lipeng@air.tsinghua.edu.cn}{\texttt{lipeng@air.tsinghua.}} \href{mailto:lipeng@air.tsinghua.edu.cn}{\texttt{edu.cn}}) and Y.Liu (\href{mailto:liuyang2011@tsinghua.edu.cn}{\texttt{liuyang2011@tsinghua.edu.cn}}).} \\
    $^1$Dept. of Comp. Sci. \& Tech., Institute for AI, Tsinghua University, Beijing, China \\
    $^1$Beijing National Research Center for Information Science and Technology \\
    $^2$Institute for AI Industry Research, Tsinghua University, Beijing, China \\
    $^3$Beijing Academy of Artificial Intelligence, Beijing, China\\
    $^4$International Innovation Center of Tsinghua University, Shanghai, China \\
    $^5$Quan Cheng Laboratory $^6$Zhongguancun Laboratory, Beijing, P.R.China $^7$Tencent AI Lab \\
}

\begin{document}
\maketitle
\begin{abstract}
Machine translation systems are expected to cope with various types of constraints in many practical scenarios. While neural machine translation (NMT) has achieved strong performance in unconstrained cases, it is non-trivial to impose pre-specified constraints into the translation process of NMT models. Although many approaches have been proposed to address this issue, most existing methods can not satisfy the following three desiderata at the same time: (1) high translation quality, (2) high match accuracy, and (3) low latency. In this work, we propose a template-based method that can yield results with high translation quality and match accuracy and the inference speed of our method is comparable with unconstrained NMT models.
Our basic idea is to rearrange the generation of constrained and unconstrained tokens through a template.
Our method does not require any changes in the model architecture and the decoding algorithm. Experimental results show that the proposed template-based approach can outperform several representative baselines in both lexically and structurally constrained translation tasks.~\footnote{The source code is available at \url{https://github.com/THUNLP-MT/Template-NMT}.}
\end{abstract}

\section{Introduction}

Constrained machine translation is of important value for a wide range of practical applications, such as interactive translation with user-specified lexical constraints~\cite{Koehn:2009:Lexical,Li:2020:Lexical,Jon:2021:Append},
domain adaptation with in-domain dictionaries~\cite{Michon:2020:Domain,Niehues:2021:Dictionaries}, 
and webpage translation with markup tags as structural constraints~\cite{Hashimoto:2019:Structure,Hanneman:2020:Structure}. Developing constrained neural machine translation (NMT) approaches can make NMT models applicable to more real-world scenarios~\cite{Toms:2021:facilitating}. 

However, it is challenging to directly impose constraints for NMT models due to their end-to-end nature~\cite{Post:2018:Lexical}.
In accordance with this problem, a branch of studies modifies the decoding algorithm to take the constraints into account when selecting candidates~\cite{Hokamp:2017:Lexical,Hasler:2018:Lexical,Post:2018:Lexical,Hu:2019:Lexical,Hashimoto:2019:Structure}.
Although constrained decoding algorithms can guarantee the presence of constrained tokens, they can significantly slow down the translation process~\cite{Wang:2022:Lexical} and can sometimes result in poor translation quality~\cite{Zhang:2021:Lexical}.

Another branch of works constructs synthetic data to help NMT models acquire the ability to translate with constraints~\cite{Song:2019:Lexical,Dinu:2019:Lexical,Michon:2020:Domain}.
For instance,
\citet{Hanneman:2020:Structure} propose to inject markup tags into plain parallel texts to learn structurally constrained NMT models. The major drawback of data augmentation based methods is that they sometimes violate the constraints~\cite{Hanneman:2020:Structure,Chen:2021:Lexical}, limiting their application in constraint-critical situations.

In this work, we use {\em free tokens} to denote the tokens that are not covered by the provided constraints.
Our motivation is to decompose the whole constrained translation task into the arrangement of constraints and the generation of free tokens. The constraints can be of many types, ranging from phrases in lexically constrained translation to markup tags in structurally constrained translation. Intuitively, only arranging the provided constraints into the proper order is much easier than generating the whole sentence.
Therefore, we build a template by abstracting free token fragments into nonterminals, which are used to record the relative position of all the involved fragments. The template can be treated as a plan of the original sentence. The arrangement of constraints can be learned through a {\em template generation} sub-task.

Once the template is generated, we need some derivation rules to convert the nonterminals mentioned above into free tokens. Each derivation rule shows the correspondence between a nonterminal and a free token fragment. These rules can be learned by the NMT model through semi-structured data.
We call this sub-task {\em template derivation}.
During inference, the model firstly generates the template and then extends each nonterminal in the template into natural language text.
Note that the two proposed sub-tasks can be accomplished through a single decoding pass. Thus the decoding speed of our method is comparable with unconstrained NMT systems.
By designing template format, our approach can cope with different types of constraints, such as lexical constraints, XML structural constraints, or Markdown constraints.

\paragraph{Contributions} In summary, the contributions of this work can be listed as follows:
\begin{itemize}
  \item We propose a novel template-based constrained translation framework to disentangle the generation of constraints and free tokens.
  \item We instantiate the proposed framework with both lexical and structural constraints, demonstrating the flexibility of this framework.
  \item Experiments show that our method can outperform several strong baselines, achieving high translation quality and match accuracy while maintaining the inference speed. 
\end{itemize}

\section{Related Work}

\subsection{Lexically Constrained Translation}

Several researchers direct their attention to modifying the decoding algorithm to impose lexical constraints~\cite{Hasler:2018:Lexical}. For instance, \citet{Hokamp:2017:Lexical} propose  grid beam search (GBS) that organizes candidates in a grid, which enumerates the provided constrained tokens at each decoding step. However, the computation complexity of GBS scales linearly with the number of constrained tokens. To reduce the runtime complexity, \citet{Post:2018:Lexical} propose dynamic beam allocation (DBA), which divides a fixed size of beam for candidates having met the same number of constraints. \citet{Hu:2019:Lexical} propose to vectorize DBA further. The resulting VDBA algorithm is still significantly slower compared with the vanilla beam search algorithm~\cite{Wang:2022:Lexical}.

Another line of studies trains the model to copy the constraints through data augmentation. \citet{Song:2019:Lexical} propose to replace the corresponding source phrases with the target constraints, and \citet{Dinu:2019:Lexical} propose to insert target constraints as inline annotations. Some other works propose to append target constraints to the whole source sentence as side constraints~\cite{Chen:2020:Lexical,Niehues:2021:Dictionaries,Jon:2021:Append}. Although these methods introduce little additional computational overhead at inference time, they can not guarantee the appearance of the constraints~\cite{Chen:2021:Lexical}. \citet{Xiao:2022:Lexical} transform constrained translation into a bilingual text-infilling task. A limitation of text-infilling is that it can not reorder the constraints, which may negatively affect the translation quality for distinct language pairs.

Recently, some researchers have tried to adapt the architecture of NMT models for this task. \citet{Susanto:2020:Lexical} adopt non-autoregressive translation models~\cite{Gu:2019:Levenshtein} to insert target constraints.
\citet{Wang:2022:Lexical} prepend vectorized keys and values to the attention modules~\cite{Vaswani:2017:Transformer} to integrate constraints. However, their model may still suffer from low match accuracy when decoding without VDBA. In this work, our method can achieve high translation quality and match accuracy without significantly increasing the inference overhead.

\subsection{Structurally Constrained Translation}

Structurally constrained translation is useful since text data is often wrapped with markup tags on the Web~\cite{Hashimoto:2019:Structure}, which is an essential source of information for humans.
Compared with lexically constrained translation, structurally constrained translation is relatively unexplored.
\citet{Joanis:2013:Structure} examine a two-stage method for statistical machine translation systems, which firstly translates the plain text and then injects the tags based on phrase alignments and some carefully designed rules.
Moving to the NMT paradigm, large-scale parallel corpora with structurally aligned markup tags are scarce. \citet{Hanneman:2020:Structure} propose to inject tags into plain text to create synthetic data. \citet{Hashimoto:2019:Structure} collect a parallel dataset consisting of structural text translated by human experts.
\citet{Zhang:2021:Lexical} propose a constrained decoding algorithm to translate structured text. However, their method significantly slows down the translation process.

In this work, our approach can be easily extended for structural constraints, leaving the decoding algorithm unchanged.
The template in our approach can be seen as an intermediate plan, which has been investigated in the field of data-to-text generation~\cite{Moryossef:2019:Step}. \citet{Zhang:2019:Neural} also explored the idea of disentangling different parts in a sentence using special tokens.

\section{Approach}

\subsection{Template-based Machine Translation}

Given a source-language sentence $\mathbf{x} = x_1 \cdots x_I$ and a target-language sentence $\mathbf{y} = y_1 \cdots y_J$, an NMT model is trained to estimate the conditional probability $P(\mathbf{y} | \mathbf{x}; \bm{\theta})$, which can be given by
\begin{equation}
  P(\mathbf{y} | \mathbf{x}; \bm{\theta}) = \prod_{j=1}^{J} P(y_j | \mathbf{x}, \mathbf{y}_{<j}; \bm{\theta}),
\end{equation}
where $\bm{\theta}$ is the set of parameters to optimize and $\mathbf{y}_{<j}$ is the partial translation at the $j$-th step.

In this work,
we firstly build a template to simplify the whole sentence.
Formally, we use $\mathbf{s}$ and $\mathbf{t}$ to represent the source- and target-side templates, respectively. In the template, free token fragments are abstracted into nonterminals. We use $\mathbf{e}$ and $\mathbf{f}$ to denote the derivation rules of the nonterminals for the source and target template, respectively.

The model is trained on two sub-tasks. 
Firstly, the model learns to generate the target template $\mathbf{t}$:
\begin{equation}
  \label{eq: task1}
  P(\mathbf{t} | \mathbf{s}, \mathbf{e}; \bm{\theta}) = \prod_{j=1}^{T}
  P(t_j | \mathbf{s}, \mathbf{e}, \mathbf{t}_{<j}; \bm{\theta}).
\end{equation}

Secondly, we train the same model to estimate the conditional probability of $\mathbf{f}$:
\begin{equation}
  \label{eq: task2}
  P(\mathbf{f} | \mathbf{s}, \mathbf{e}, \mathbf{t}; \bm{\theta}) = \prod_{j=1}^{F}
  P(f_j | \mathbf{s}, \mathbf{e}, \mathbf{t}, \mathbf{f}_{<j}; \bm{\theta}).
\end{equation}

The target sentence $\mathbf{y}$ can be reconstructed by extending each nonterminal in $\mathbf{t}$ using the corresponding derivation rule in $\mathbf{f}$.
We can jointly learn the two sub-tasks in one pass to improve both the training and inference efficiency. Formally, the model is trained to maximize the following joint probability of $\mathbf{t}$ and $\mathbf{f}$ in practice:
\begin{equation}
\label{eq: two tasks}
\begin{split}
  P(\mathbf{t}, \mathbf{f} | \mathbf{s}, \mathbf{e}; \bm{\theta}) 
  = P(\mathbf{t} | \mathbf{s}, \mathbf{e}; \bm{\theta}) \times
  P(\mathbf{f} | \mathbf{s}, \mathbf{e}, \mathbf{t}; \bm{\theta}).
\end{split}
\end{equation}

\begin{figure*}[ht]
  \centering
  \includegraphics[width=0.995\textwidth]{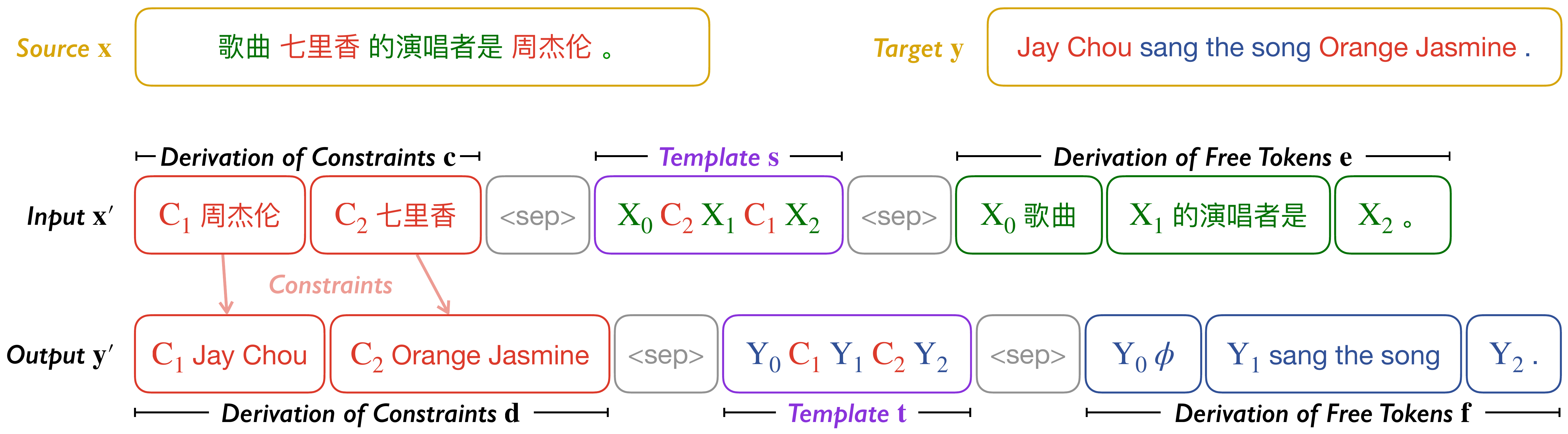}
  \caption{
      Example for lexically constrained translation. The constraints are $\langle${\color{sred}\begin{CJK}{UTF8}{gbsn}周杰伦\end{CJK}}, {\color{sred}Jay Chou}$\rangle$ and $\langle${\color{sred}\begin{CJK}{UTF8}{gbsn}七里香\end{CJK}}, {\color{sred}Orange Jasmine}$\rangle$.
      Note that {\color{sgreen}$\mathrm{X}_n$} and {\color{sblue}$\mathrm{Y}_n$} are \textbf{not} linked nonterminals, since the source and target free token fragments are not necessarily aligned.
      The derivation rule {\color{sgreen}$\mathrm{X}_0 \rightarrow$ \begin{CJK}{UTF8}{gbsn}歌曲\end{CJK}} is learned through the concatenation of {\color{sgreen}$\mathrm{X}_0$} and {\color{sgreen}\begin{CJK}{UTF8}{gbsn}歌曲\end{CJK}} (i.e., {\color{sgreen}$\mathrm{X}_0$ \begin{CJK}{UTF8}{gbsn}歌曲\end{CJK}}).
      ``$\phi$'' denotes an empty string.
      See Section~\ref{subsec:lexical template} for more details.
  }
  \label{fig:lexical_template}
\end{figure*}

\subsection{Template for Lexical Constraints}
\label{subsec:lexical template}

In lexically constrained translation, some source phrases in the input sentence are required to be translated into pre-specified target phrases. For a source sentence $\mathbf{x}$, we use $\left \{ \left \langle \mathbf{u}^{(n)}, \mathbf{v}^{(n)} \right \rangle \right \}_{n=1}^{N}$ to denote the given constraint pairs, where $\mathbf{u}^{(n)}$ is the $n$-th source constraint, and $\mathbf{v}^{(n)}$ is the corresponding target constraint. All the $N$ source constraints can divide $\mathbf{x}$ into $2N+1$ fragments:
\begin{equation}
  \mathbf{x} = \mathbf{p}^{(0)}\mathbf{u}^{(1)}\mathbf{p}^{(1)}\cdots
  \mathbf{u}^{(N)}\mathbf{p}^{(N)},
\end{equation}
where $\mathbf{p}^{(n)}$ is the $n$-th free token fragment. We can set $\mathbf{p}^{(0)}$ to an empty string to represent sentences that start with a constraint, and set $\mathbf{p}^{(N)}$ to an empty string for sentences that end with a constraint. We can also set $\mathbf{p}^{(n)}$ to an empty string for the cases where $\mathbf{u}^{(n)}$ and $\mathbf{u}^{(n+1)}$ are adjacent in $\mathbf{x}$. Similarly, the target sentence can be represented by
\begin{equation}
  \mathbf{y} = \mathbf{q}^{(0)}\mathbf{v}^{(i_1)}\mathbf{q}^{(1)}\cdots
  \mathbf{v}^{(i_N)}\mathbf{q}^{(N)},
\end{equation}
where $\mathbf{q}^{(n)}$ is the $n$-th free token fragment in the target sentence $\mathbf{y}$. We use $i_1,\cdots,i_N$ to denote the order of the constraints in $\mathbf{y}$. The $n$-th index $i_n$ is not necessarily equal to $n$, since the order of the constraints in the target sentence $\mathbf{y}$ is often different from that in the source sentence $\mathbf{x}$. 

We then abstract each fragment of text into nonterminals to build the template for lexically constrained translation. Concretely, the $n$-th free token fragment in the source sentence $\mathbf{x}$ is abstracted into $\mathrm{X}_n$, for each $n \in \{0, \cdots, N\}$. The $n$-th free token fragment in the target sentence is abstracted into $\mathrm{Y}_n$, for each $n \in \{0, \cdots, N\}$. In order to indicate the alignment between corresponding source and target constraints, we abstract $\mathbf{u}_n$ and $\mathbf{v}_n$ into the same nonterminal $\mathrm{C}_n$. Note that $\mathrm{X}_n$ and $\mathrm{Y}_n$ are \textbf{not} linked nonterminals,
since fragments of free tokens are not bilingually aligned. The resulting source- and target-side templates are given by
\begin{equation}
\begin{split}
  \mathbf{s} &= \mathrm{X}_0\mathrm{C}_1\mathrm{X}_1\cdots\mathrm{C}_N\mathrm{X}_N, \\
  \mathbf{t} &= \mathrm{Y}_0\mathrm{C}_{i_1}\mathrm{Y}_1\cdots\mathrm{C}_{i_N}\mathrm{Y}_N. \\
\end{split}
\end{equation}

We need to define some derivation rules to convert the template into a natural language sentence. The derivation of nonterminals can be seen as the inverse of the abstraction process. Thus the derivation of the target-side template $\mathbf{t}$ would be
\begin{equation}
  \begin{split}
  \mathrm{C}_n &\rightarrow \mathbf{v}^{(n)}\quad \text{for\ each}\ \ n \in \{1, \cdots, N\},\\
  \mathrm{Y}_n &\rightarrow \mathbf{q}^{(n)}\quad \text{for\ each}\ \ n \in \{0, \cdots, N\}.\\
\end{split}
\end{equation}

The derivation of the source-side template $\mathbf{s}$ can be defined similarly.
Note that $\mathrm{C}_n$ produces the $n$-th source constraint $\mathbf{u}_n$ at the source side while producing the target constraint $\mathbf{v}_n$ at the target side.
In order to make the derivation rules learnable by NMT models, we propose to use the concatenation of the nonterminal and the corresponding sequence of terminals to denote each derivation rule.
For example, we use $\mathrm{Y}_n \mathbf{q}^{(n)}$ to represent $\mathrm{Y}_n \rightarrow \mathbf{q}^{(n)}$. We use $\mathbf{d}$ and $\mathbf{f}$ to denote the derivation of constraints and free tokens at the target side, respectively:
\begin{equation}
\begin{split}
  \mathbf{d} &= \mathrm{C}_1 \mathbf{v}^{(1)} \cdots \mathrm{C}_N \mathbf{v}^{(N)}, \\
  \mathbf{f} &= \mathrm{Y}_0 \mathbf{q}^{(0)} \cdots \mathrm{Y}_N \mathbf{q}^{(N)}.
\end{split}
\end{equation}

At the source side, we use $\mathbf{c}$ and $\mathbf{e}$ to denote the derivation of constraints and free tokens, respectively. $\mathbf{c}$ and $\mathbf{e}$ can be defined similarly.
Since the constraints are pre-specified by the users, the model only needs to learn the derivation of free tokens. To this end, we place the derivation of constraint-related nonterminals before the template as a conditional prefix. Then the model learns the generation of the template and the derivation of free tokens, step by step.

The final format of the input and output sequences at training time can be given by
\begin{equation}
\begin{split}
  \mathbf{x}^{\prime} &= \mathbf{c}\ \texttt{<sep>}\ \mathbf{s}\ \texttt{<sep>}\ \mathbf{e},\\
  \mathbf{y}^{\prime} &= \mathbf{d}\ \texttt{<sep>}\ \mathbf{t}\ \texttt{<sep>}\ \mathbf{f},
\end{split}
\end{equation}
respectively.
We use the delimiter {\tt <sep>} to separate the template and the derivations.
Figure~\ref{fig:lexical_template} gives an example of both $\mathbf{x}^{\prime}$ and $\mathbf{y}^{\prime}$.
At inference time, we feed $\mathbf{x}^{\prime}$ to the encoder, and provide ``$\mathbf{d}$ {\tt <sep>}'' to the decoder as the constrained prefix. Then the model generates the remaining part of $\mathbf{y}^{\prime}$ (i.e., ``$\mathbf{t}$ {\tt <sep>} $\mathbf{f}$'').

\begin{figure}[t]
  \centering
  \includegraphics[width=0.42\textwidth]{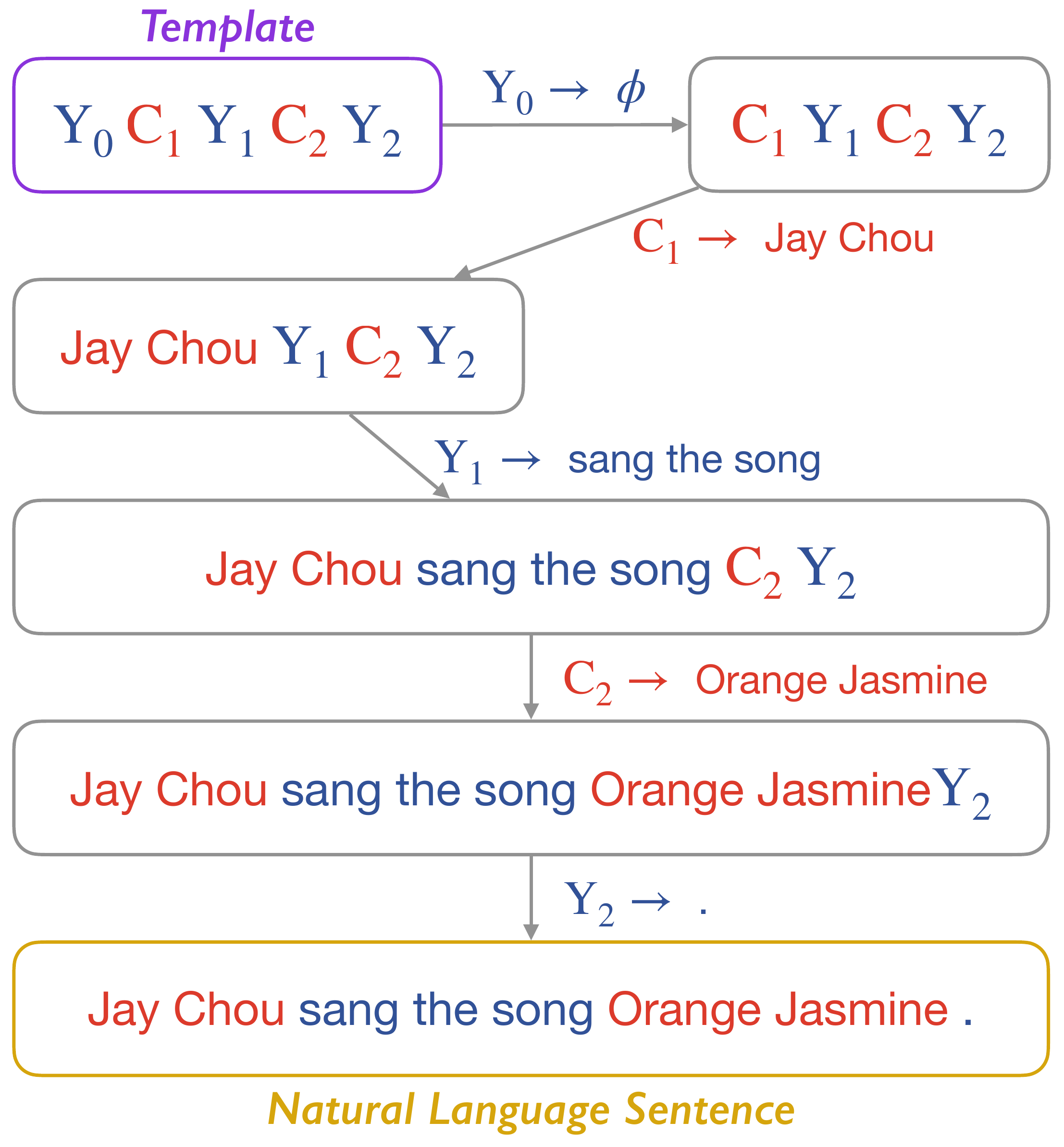}
  \caption{
      The template can be converted into a natural language sentence by replacing the nonterminals according to the corresponding derivation rules.
  }
  \label{fig:template_derivation}
\end{figure}

Figure~\ref{fig:template_derivation} explains the way we convert the output sequence into a natural language sentence. The conversion from the template to the target-language sentence can be done through a simple script, and the computational cost caused by the conversion is negligible, compared with the model inference.

\begin{figure*}[ht]
  \centering
  \includegraphics[width=0.995\textwidth]{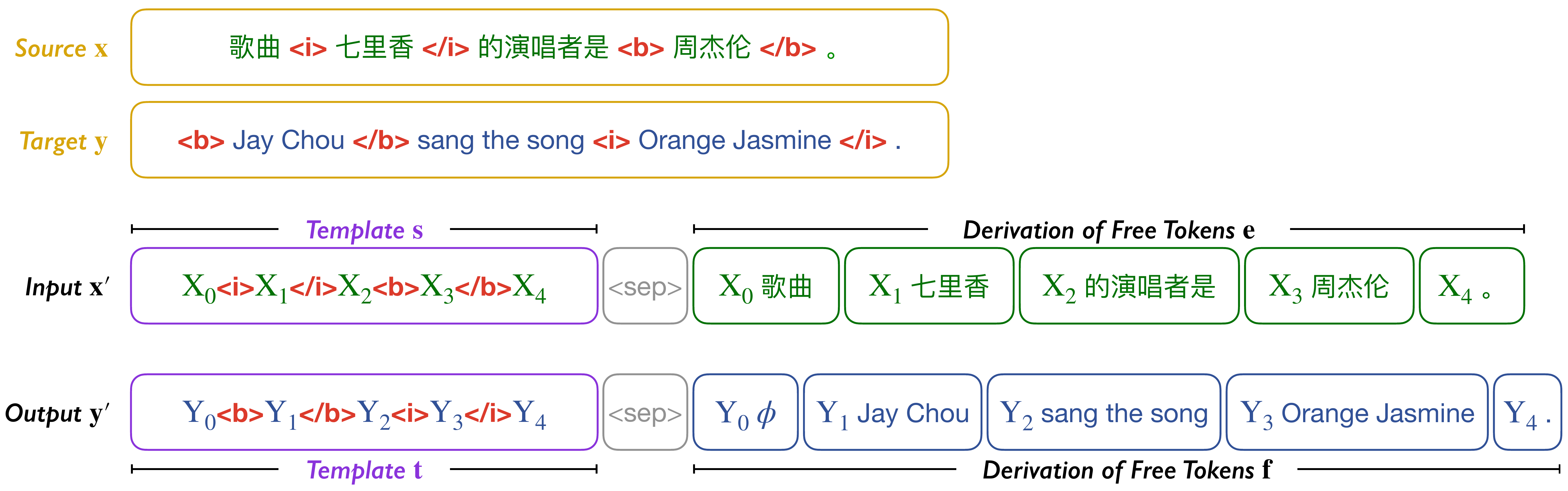}
  \caption{
      Example for structurally constrained translation. The markup tags are reserved in the template, while free tokens are abstracted.
      Note that {\color{sgreen}$\mathrm{X}_n$} and {\color{sblue}$\mathrm{Y}_n$} are \textbf{not} linked nonterminals.
      See Section~\ref{subsec:structural template} for more details.
  }
  \label{fig:structural_template}
\end{figure*}

Note that we also abstract the constraints when building the template. The reason is that the model only needs to generate the order of constraints in this way, rather than copy all the specific tokens, which may suffer from copy failure~\cite{Chen:2021:Lexical}.
The formal representation for our lexically constrained model is slightly different from that defined in Eq.~(\ref{eq: two tasks}), which should be changed into
\begin{equation}
\begin{split}
  &P(\mathbf{t}, \mathbf{f} | \mathbf{c}, \mathbf{s}, \mathbf{e}, \mathbf{d}; \bm{\theta}) \\
  = &P(\mathbf{t} | \mathbf{c}, \mathbf{s}, \mathbf{e}, \mathbf{d}; \bm{\theta}) \times
  P(\mathbf{f} | \mathbf{c}, \mathbf{s}, \mathbf{e}, \mathbf{d}, \mathbf{t}; \bm{\theta}).
\end{split}
\end{equation}

\subsection{Template for Structural Constraints}
\label{subsec:structural template}

The major challenge of structured text translation is to maintain the correctness of the structure, which is often indicated by markup tags~\cite{Hashimoto:2019:Structure}. 
The proposed framework can also deal with structurally constrained translation. Similarly, we replace free token fragments with nonterminals to build the template, where the markup tags are reserved. Figure~\ref{fig:structural_template} shows an example. Formally, given a sentence pair $\langle \mathbf{x}, \mathbf{y} \rangle$ with $N$ markup tags, the source- and target-side templates are given by
\begin{equation}
\begin{split}
  \mathbf{s} &= \mathrm{X}_0\texttt{<tag$_{1}$>}\mathrm{X}_1\cdots\texttt{<tag$_{N}$>}\mathrm{X}_N, \\
  \mathbf{t} &= \mathrm{Y}_0\texttt{<tag$_{i_1}$>}\mathrm{Y}_1\cdots\texttt{<tag$_{i_N}$>}\mathrm{Y}_N, \\
\end{split} 
\end{equation}
respectively.
The order of markup tags at the target side (i.e., $i_1 \cdots i_N$) may be different from that at the source side (i.e., $1 \cdots N$).

For each $n \in \{0, \cdots, N\}$, $\mathrm{X}_n$ can be derived into the $n$-th source-side free token fragment $\mathbf{p}^{(n)}$, and $\mathrm{Y}_n$ can be extended into the target-side free token fragment $\mathbf{q}^{(n)}$.
$\mathrm{X}_n$ and $\mathrm{Y}_n$ are \textbf{not} linked.
The derivation sequences can be defined as
\begin{equation}
  \begin{split}
    \mathbf{e} &= \mathrm{X}_0 \mathbf{p}^{(0)} \cdots \mathrm{X}_N \mathbf{p}^{(N)}, \\
    \mathbf{f} &= \mathrm{Y}_0 \mathbf{q}^{(0)} \cdots \mathrm{Y}_N \mathbf{q}^{(N)}.
\end{split} 
\end{equation}

The format of the input and output would be
\begin{equation}
\begin{split}
  \mathbf{x}^{\prime} &= \mathbf{s}\ \texttt{<sep>}\ \mathbf{e},\\
  \mathbf{y}^{\prime} &= \mathbf{t}\ \texttt{<sep>}\ \mathbf{f},
\end{split}
\end{equation}
respectively. Figure~\ref{fig:structural_template} illustrates an example for both $\mathbf{x}^{\prime}$ and $\mathbf{y}^{\prime}$. The formal representation of our structurally constrained model is the same as Eq.~(\ref{eq: two tasks}). The model arranges the markup tags when generating $\mathbf{t}$ and completes the whole sentence when generating $\mathbf{f}$, which is consistent with our motivation to decompose the whole task into constraint arrangement and free token generation.

\section{Lexically Constrained Translation}
\subsection{Setup}
\label{subsec:lexical-setup}
\paragraph{Parallel Data} We conduct experiments on two language pairs, including English-Chinese and English-German. For English-Chinese, we use the dataset of WMT17 as the training corpus, consisting of 20.6M sentence pairs. For English-German, the training data is from WMT20, containing 41.0M sentence pairs.
We provide more details of data preprocessing in Appendix.
Following recent studies on lexically constrained translation~\cite{Chen:2021:Lexical,Wang:2022:Lexical}, we evaluate our method on human-annotated alignment test sets. For English-Chinese, both the validation and test sets are from \citet{Liu:2005:Align}. For English-German, the test set is from \citet{Zenkel:2020:Align}. We use {\tt newstest2013} as the validation set, whose word alignment is annotated by {\tt fast-align}\footnote{\url{https://github.com/clab/fast_align}}. The training sets are filtered to exclude test and validation sentences.

\paragraph{Lexical Constraints} Following some recent works~\cite{Song:2019:Lexical,Chen:2020:Lexical,Chen:2021:Lexical,Wang:2022:Lexical}, we simulate real-world lexically constrained translation scenarios by sampling constraints from the phrase table that are extracted from parallel sentence pairs based on word alignment. The script used to create the constraints is publicly available.\footnote{\url{https://github.com/ghchen18/cdalign/blob/main/scripts/extract_phrase.py}} Specifically, the number of constraints for each sentence pair ranges between 0 and 3, and the length of each constraint ranges between 1 and 3 tokens.
We use {\tt fast-align} to build the alignment of the training data.

\paragraph{Model Configuration} We adopt Transformer~\cite{Vaswani:2017:Transformer} as our NMT model, which is optimized by Adam~\cite{Kingma:2015:CoRR} with $\beta_{1}=0.9$, $\beta_{2}=0.98$ and $\epsilon=10^{-9}$.
Please refer to Appendix for more details on the model configuration and the training process.

\paragraph{Baselines} We compare our approach with the following six representative baselines:
\begin{itemize}
  \item Placeholder~\cite{Crego:2016:Placeholder}: replacing constrained terms with placeholders;
  \item VDBA~\cite{Hu:2019:Lexical}: modifying beam search to incorporate target-side constraints;
  \item Replace~\cite{Song:2019:Lexical}: replacing source text with the corresponding target constraints;
  \item CDAlign~\cite{Chen:2021:Lexical}: inserting target constraints based on word alignment;
  \item AttnVector~\cite{Wang:2022:Lexical}: using attention keys and values to model constraints;
  \item TextInfill~\cite{Xiao:2022:Lexical}: filling free tokens through a bilingual text-infilling task.
\end{itemize}

\paragraph{Evaluation Metrics} We follow \citet{Alam:2021:TermEval} to use the following four metrics to make a thorough comparison of the involved methods:
\begin{itemize}
  \item BLEU~\cite{Papineni:2001:BLEU}: measuring the translation quality of the whole sentence;
  \item Exact Match: indicating the accuracy that the source constraints in the input sentences are translated into the provided target constraints;
  \item Window Overlap: quantifying the overlap ratio between the hypothesis and the reference windows for each matched target constraint, indicating if this constraint is placed in a suitable context. The window size is set to 2.
  \item 1-TERm: modifying TER~\cite{Snover:2006:TER} by setting the edit cost of constrained tokens to 2 and the cost of free tokens to 1.
\end{itemize}

We use sacreBLEU\footnote{English-Chinese: nrefs:1 | case:mixed | eff:no | tok:zh | smooth:exp | version:2.0.0. English-German: nrefs:1 | case:mixed | eff:no | tok:13a | smooth:exp | version:2.0.0.}~\cite{Post:2018:BLEU} to estimate the BLEU score, and adapt the scripts released by \citet{Alam:2021:TermEval} for the other three metrics.

\begin{table*}[ht]
  \centering
  \small
  \scalebox{0.92}{
  \begin{tabular}{l m{14.5cm}}
  \toprule
  \bf Constraints & \begin{CJK}{UTF8}{gbsn}{\normalsize $\langle${\color{sred}slowing down},{\color{sred}减弱}$\rangle$}; {\normalsize$\langle${\color{sblue}price hike},{\color{sblue}价格上涨}$\rangle$}\end{CJK}\\
  \cmidrule(lr){1-1} \cmidrule(lr){2-2}
  \bf Source & {Analysts are concerned that since there is no sign yet of any }{\normalsize\color{sred}slowing down} {\small of this} {\normalsize\color{sblue}price hike} {, the prospect of the British real estate market as where it is heading now is far from optimistic.}\\
  \cmidrule(lr){1-1} \cmidrule(lr){2-2}
  \bf Reference & \begin{CJK}{UTF8}{gbsn}{分析家担心, 由于目前还看不见}{\normalsize\color{sblue}\ 价格上涨 }{趋势有}{\normalsize\color{sred}\ 减弱 }{的迹象, 照此发展下去, 英国房地产市场前景堪忧。}\end{CJK}\\
  \cmidrule(lr){1-1} \cmidrule(lr){2-2}
  {\bf Input} (enc) & {\normalsize\color{sred}$\mathrm{C}_1$ slowing down} {\normalsize\color{sblue}$\mathrm{C}_2$ price hike} {\tt <sep>} $\mathrm{X}_0$ {\normalsize\color{sred}$\mathrm{C}_1$} $\mathrm{X}_1$ {\normalsize\color{sblue}$\mathrm{C}_2$} $\mathrm{X}_2$ {\tt <sep>} $\mathrm{X}_0$ {Analysts are concerned that since there is no sign yet of any} $\mathrm{X}_1$ { of this} $\mathrm{X}_2$ {, the prospect of the British real estate market as where it is heading now is far from optimistic.} \\
  \cmidrule(lr){1-1} \cmidrule(lr){2-2}
  {\bf Prefix} (dec) & \begin{CJK}{UTF8}{gbsn}{\normalsize\color{sred}$\mathrm{C}_1$ 减弱} {\normalsize\color{sblue}$\mathrm{C}_2$ 价格上涨} {\tt <sep>}\end{CJK} \\
  \cmidrule(lr){1-1} \cmidrule(lr){2-2}
  \bf Output & \begin{CJK}{UTF8}{gbsn} $\mathrm{Y}_0$ {\normalsize\color{sblue}$\mathrm{C}_2$} $\mathrm{Y}_1$ {\normalsize\color{sred}$\mathrm{C}_1$} $\mathrm{Y}_2$ {\tt <sep>} $\mathrm{Y}_0$ { 分析师们担心, 由于目前还没有迹象显示} $\mathrm{Y}_1$ { 会} $\mathrm{Y}_2$ {\small, 英国房地产市场的前景远不乐观。} \end{CJK}\\
  \cmidrule(lr){1-1} \cmidrule(lr){2-2}
  \bf Result & \begin{CJK}{UTF8}{gbsn}{分析师们担心, 由于目前还没有迹象显示}{\normalsize\color{sblue}\ 价格上涨 }{会}{\normalsize \color{sred}\ 减弱 }{, 英国房地产市场的前景远不乐观。}\end{CJK}\\
  \bottomrule
  \end{tabular}}
  \caption{An example of our method. We replace the nonterminals in the template using the derivation rules to reconstruct the final result (i.e., ``\textbf{Result}'').
  Surprisingly, we find that our model can automatically sort the provided constraints when generating the template. In this example, {\color{sred}$\mathrm{C}_1$} is before {\color{sblue}$\mathrm{C}_2$} in the source-side template. But in the target-side template generated by our model, {\color{sblue}$\mathrm{C}_2$} is before {\color{sred}$\mathrm{C}_1$}, which is more suitable for the target language.}
  \label{tab:lexical-case}
\end{table*}
\begin{table*}[ht]
  \centering
  \small
  \begin{tabular}{l m{3pt} rrrr m{3pt} rrrr}
    \toprule
    \multirow{2}{*}{\bf Method} && \multirow{2}{*}{\bf BLEU} & \bf Exact & \bf Window & \multirow{2}{*}{\bf 1-TERm} && \multirow{2}{*}{\bf BLEU} & \bf Exact & \bf Window & \multirow{2}{*}{\bf 1-TERm} \\
    && & \bf Match & \bf Overlap & && & \bf Match & \bf Overlap & \\
    \midrule
    \em Direction && \multicolumn{4}{c}{\em English-Chinese} && \multicolumn{4}{c}{\em English-German} \\
    \cmidrule(lr){1-1} \cmidrule(lr){3-6} \cmidrule(lr){8-11}
    Vanilla && 42.7 & 10.1 & 4.8 & 35.7 && 24.8 & 10.0 & 8.1 & 39.2 \\
    \cmidrule(lr){1-1} \cmidrule(lr){3-6} \cmidrule(lr){8-11}
    Placeholder && 46.6 & 99.4 & 33.9 & 41.5 && 27.2 & \bf 100.0 & 29.4 & 44.6 \\
    VDBA && 45.8 & 99.6 & 33.4 & 41.7 && 29.0 & \bf 100.0 & 31.1 & 45.1 \\
    Replace && 46.4 & 93.8 & 35.5 & 40.7 && 31.1 & 96.6 & 35.7 & \underline{48.3} \\
    CDAlign && 46.2 & 92.1 & 31.7 & 41.6 && 29.7 & 95.9 & 32.3 & 46.3 \\
    AttnVector && \underline{46.9} & 93.8 & \underline{35.8} & \underline{42.4} && \underline{31.3} & 97.5 & \underline{37.2} & 47.9 \\
    TextInfill && 45.6 & \bf 100.0 & 32.8 & 39.9 && 30.7 & \bf 100.0 & 35.5 & 47.1 \\
    \cmidrule(lr){1-1} \cmidrule(lr){3-6} \cmidrule(lr){8-11}
    Ours && \bf 47.5 & \bf{100.0} & \bf{36.9} & \bf{43.1} && \bf 32.3 & \bf 100.0 & \bf 38.5 & \bf 49.8 \\
    \bottomrule
  \end{tabular}
  \caption{Results of the lexically constrained translation task for both English-Chinese and English-German. For clarity, we highlight the {\bf highest} score in bold and the \underline{second-highest} score with underlines.}
  \label{tab:lexical-main}
\end{table*}

\subsection{Main Results}
\label{sec:lexical-main-res}

\paragraph{Template Accuracy} We firstly examine the performance of the model in the template generation sub-task before investigating the translation performance.
We compare the target-side template extracted from the reference sentence and the one generated by the model to calculate the accuracy of template generation.
Formally, if the reference template $\mathbf{t}$ is $\mathrm{Y}_0\mathrm{C}_{i_1}\mathrm{Y}_1\cdots\mathrm{C}_{i_N}\mathrm{Y}_N$, the generated template $\hat{\mathbf{t}}$ is correct if 
\begin{itemize}
  \item $\hat{\mathbf{t}}=\mathrm{Y}_0\mathrm{C}_{j_1}\mathrm{Y}_1\cdots\mathrm{C}_{j_N}\mathrm{Y}_N$;
  \item the set $\{j_1,\cdots,j_N\}$ equals $\{i_1,\cdots,i_N\}$.
\end{itemize}

In other words, the model must generate all the nonterminals to guarantee the presence of the provided constraints. However, the order of constraint-related nonterminals can be flexible since there often exist various suitable orders for the provided constraints. In both English-Chinese and English-German, the template accuracy of our model is 100\%. An interesting finding is that our model learns to reorder the constraints according to the style of the target language. We provide an example of constraint reordering in Table~\ref{tab:lexical-case}.

When generating the free token derivation $\mathbf{f}$, the model can recall all the nonterminals (i.e., $\mathrm{Y}_n$) presented in the template $\mathbf{t}$ in English-Chinese. In English-German, however, the model omits one free token nonterminal, of which the frequency is 0.2\%. We use empty strings for the omitted nonterminals when reconstructing the output sentence.

\paragraph{Translation Performance}
Table~\ref{tab:lexical-main} shows the results of lexically constrained translation,
demonstrating that all the investigated methods can recall more provided constraints than the unconstrained Transformer model.
Our approach can improve the BLEU score over the involved baselines. This improvement potentially comes from two aspects: (1) our system outputs can match more pre-specified constraints compared to some baselines, such as AttnVector~\cite{Wang:2022:Lexical}
(100\% vs. 93.8\%)
; (2) our method can place more constraints in appropriate context, which can be measured by window overlap.
The exact match accuracy of VDBA~\cite{Hu:2019:Lexical} is lower than 100\% due to the out-of-vocabulary problem in English-Chinese.

TextInfill~\cite{Xiao:2022:Lexical} and our approach can achieve 100\% exact match accuracy in both the two language pairs. However, TextInfill can only place the constraints in the pre-specified order, while our approach can automatically reorder the constraints. As a result, the window overlap score of our approach is higher than TextInfill.
Please refer to Table~\ref{tab:lexical-case-appendix} in Appendix for more translation examples of both our method and some baselines

\subsection{Unconstrained Translation}

A concern for lexically constrained translation methods is that they may cause poor translation quality in unconstrained translation scenarios. We thus evaluate our approach in the standard translation task, where the model is only provided with the source sentence $\mathbf{x}$. 
Under this circumstance, the input and output can be given by
\begin{equation}
\begin{split}
  \mathbf{x}^{\prime} &= \phi\ \texttt{<sep>}\ \mathrm{X}_0\ \texttt{<sep>}\ \mathrm{X}_0 \mathbf{x}, \\
  \mathbf{y}^{\prime} &= \phi\ \texttt{<sep>}\ \mathrm{Y}_0\ \texttt{<sep>}\ \mathrm{Y}_0 \mathbf{y}, \\
\end{split}
\end{equation}
respectively. The BLEU scores of our method are 42.6 and 25.0 for English-Chinese and English-German, respectively. The performance of our method is comparable with the vanilla model, which can dispel the concern that our approach may worsen the unconstrained translation quality.


\subsection{Inference Speed}

\begin{table}[ht]
  \centering
  \small
  \begin{tabular}{l c}
    \toprule
    \bf Methods & \textbf{Speed} \\
    \midrule
    Vanilla & 3392 { tokens per second}  \\
    \cmidrule(lr){1-1} \cmidrule(lr){2-2}
    Ours & 3390 { tokens per second}\\
    \bottomrule
  \end{tabular}
  \caption{Inference speed of our method and the vanilla model on the English-Chinese validation set.}
  \label{tab:speed}
\end{table}

Table~\ref{tab:speed} shows the decoding speed. Since we did not change the model architecture and the decoding algorithm, the speed of our method is close to the vanilla Transformer model~\cite{Vaswani:2017:Transformer}. Although our speed is almost the same as the vanilla model, our inference time is a bit longer, given the fact that the output sequence $\mathbf{y}^{\prime}$ is longer than the original target-language sentence $\mathbf{y}$.

\begin{table*}[ht]
    \centering
    \small
    \begin{tabular}{l m{3pt} rrr m{3pt} rrr}
      \toprule
      \multirow{2}{*}{\bf Method} && \multirow{2}{*}{\bf BLEU} & \multicolumn{2}{c}{\bf Structure Accuracy} && \multirow{2}{*}{\bf BLEU} & \multicolumn{2}{c}{\bf Structure Accuracy} \\
      \cmidrule(lr){4-5} \cmidrule(lr){8-9}
      && & \bf Correct & \bf Match && & \bf Correct & \bf Match \\
      \midrule
      \em Direction && \multicolumn{3}{c}{\em English-French} && \multicolumn{3}{c}{\em English-Russian} \\
      \cmidrule(lr){1-1} \cmidrule(lr){3-5} \cmidrule(lr){7-9}
      Remove && 31.4 & n/a & n/a && 21.0 & n/a & n/a \\
      Split-Inject && \underline{66.1} & \bf 100.00 & \bf 100.00 && 43.1 & \bf 100.00 & \bf 99.85 \\
      XML && 65.3 & 99.55 & 99.30 && \underline{44.9} & 99.45 & 98.90 \\
      \cmidrule(lr){1-1} \cmidrule(lr){3-5} \cmidrule(lr){7-9}
      Ours && \bf 67.3 & \bf 100.00 & \bf 100.00 && \bf 45.8 & \bf 100.00 & \underline{99.80} \\
      \midrule
      \em Direction && \multicolumn{3}{c}{\em English-Chinese} && \multicolumn{3}{c}{\em English-German} \\
      \cmidrule(lr){1-1} \cmidrule(lr){3-5} \cmidrule(lr){7-9}
      Remove && 31.5 & n/a & n/a && 25.7 & n/a & n/a \\
      Split-Inject && 57.0 & \bf 100.00 & 99.30 && 50.7 & \bf 100.00 & \bf 99.80 \\
      XML && \underline{61.2} & 99.85 & \underline{99.75} && \underline{52.7} & 99.80 & 99.20 \\
      \cmidrule(lr){1-1} \cmidrule(lr){3-5} \cmidrule(lr){7-9}
      Ours && \bf 61.5 & \bf 100.00 & \bf 99.80 && \bf 53.6 & \bf 100.00 & \bf 99.80 \\
      \bottomrule
    \end{tabular}
    \caption{Results of the structurally constrained translation task. We highlight the {\bf highest} score in bold and the \underline{second-highest} score with underlines.}
    \label{tab:structure-main}
  \end{table*}

  \subsection{Effect of Data Scale}

  \begin{figure}[ht]
    \centering
    \includegraphics[width=0.48\textwidth]{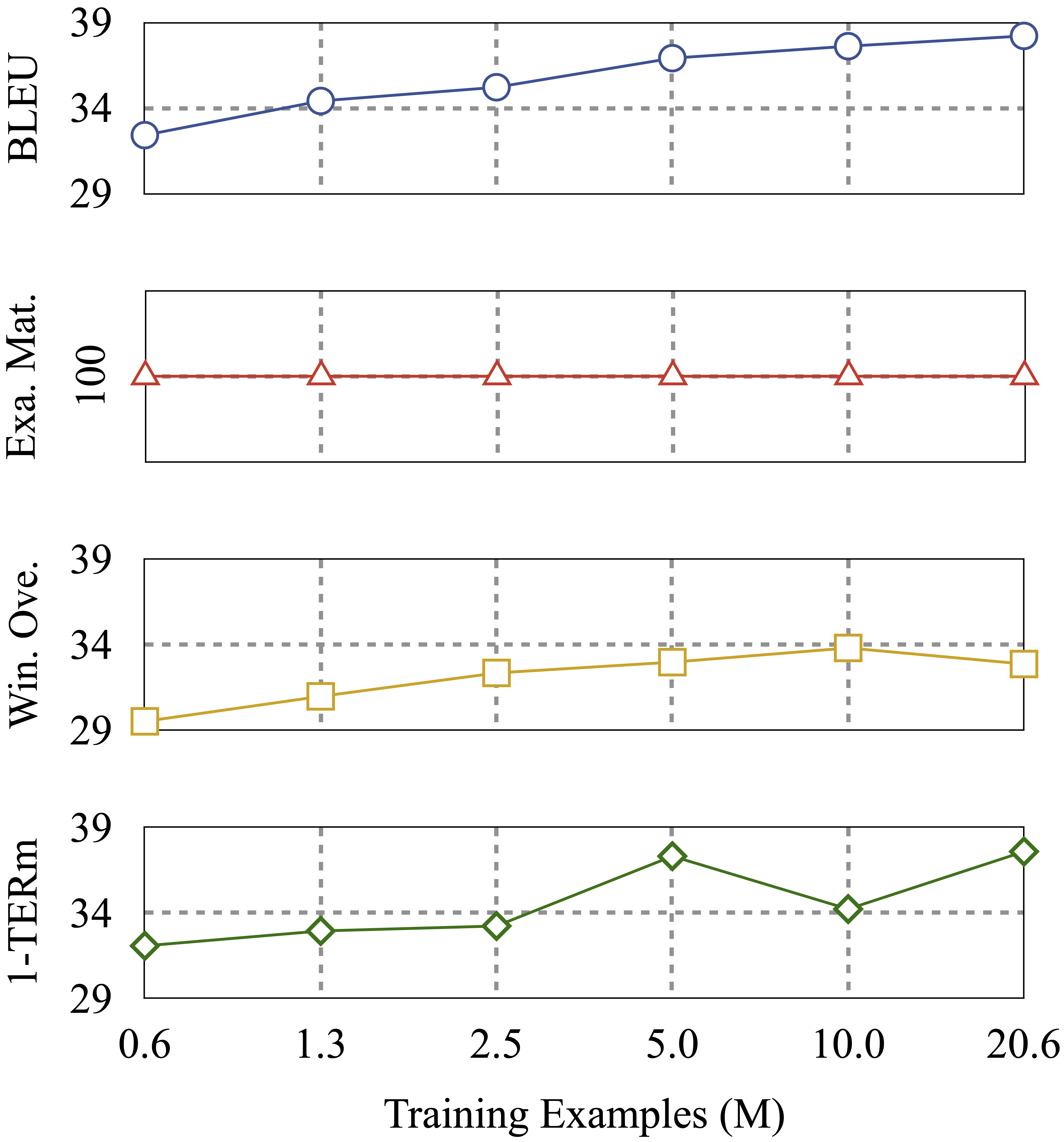}
    \caption{
        Effect of data scale. The results are reported on the English-Chinese validation set.
    }
    \label{fig:data-scale}
  \end{figure}
  
  We vary the amounts of training data to investigate the effect of data scale on our approach. Figure~\ref{fig:data-scale} shows the results. The BLEU score increases with the data size, while the window overlap score reaches the highest value when using 10.0M training examples. When using all the training data, the 1 - TERm metric achieves the best value. We find that the exact match accuracy of our method is maintained at 100\%, even with only 0.6M training examples. This trend implies that our method can be applied in some low-resource scenarios.

\subsection{More Analysis}

Due to space limitation, we place a more detailed analysis of our approach in Appendix, including the effect of the alignment model, the performance on more language pairs, and the domain robustness of our model, which is evaluated on the WMT21 terminology translation task~\cite{Alam:2021:WMT21Term}
that lies in the COVID-19 domain.

\section{Structurally Constrained Translation}

\subsection{Setup}

\paragraph{Data} We conduct our experiments on the dataset released by \citet{Hashimoto:2019:Structure}, which supports the translation from English to seven other languages. We select four languages, including French, Russian, Chinese, and German. For each language pair, the training set contains roughly 100K sentence pairs. We report the results on the validation sets since the test sets are not open-sourced. 
We follow \citet{Hashimoto:2019:Structure} to use {\tt SentencePiece}\footnote{\url{https://github.com/google/sentencepiece}} to preprocess the data, which supports user-defined special symbols. The model type of {\tt SentencePiece} is set to {\tt unigram}, and the vocabulary size is set to 9000.
For English-Chinese, we over-sample the English sentences when learning the joint tokenizer, since Chinese has more unique characters than English~\cite{Hashimoto:2019:Structure}. We did not perform over-sampling for other language pairs.
We register the XML tags and URL placeholders as user-defined special symbols. In addition, we also register {\tt \&amp;}, {\tt \&lt;}, and {\tt \&gt;} as special tokens, following \citet{Hashimoto:2019:Structure}.


\paragraph{Model Configuration} Since the data scale for structurally constrained translation is much smaller than lexically constrained translation, we follow \citet{Hashimoto:2019:Structure} to set the width of the model to 256 and the depth of the model to 6. See Section~\ref{subsec:structure-model-appendix} in Appendix for more details.

\paragraph{Baselines} We compare our approach with the following three baselines:
\begin{itemize}
  \item Remove: removing the markup tags and only translating the plain text;
  \item Split-Inject~\cite{Al:1997:Split}: splitting the input sentence based on the markup tags and then translating each text fragment independently, and finally injecting the tags;
  \item XML~\cite{Hashimoto:2019:Structure}: directly learning the NMT model end-to-end using parallel sentences with XML tags.
\end{itemize}

\paragraph{Evaluation Metrics} We follow \citet{Hashimoto:2019:Structure} to use the following metrics:
\begin{itemize}
  \item BLEU: considering the structure when estimating BLEU score~\cite{Papineni:2001:BLEU};
  \item Structure Accuracy: utilizing the {\tt etree} package to check if the system output is a valid XML structure (i.e., Correct), and if the output structure exactly matches the structure of the given reference (i.e., Match). 
\end{itemize}

All the metrics are calculated using the evaluation script released by \citet{Hashimoto:2019:Structure}.

\subsection{Main Results}

\paragraph{Template Accuracy}
We firstly examine the accuracy of the generated templates. A generated template is correct if
\begin{itemize}
    \item the template is a valid XML structure;
    \item the template recalls all the markup tags of the input sentence.
\end{itemize}

The template accuracy of our method is 100\% in all the four language pairs. Similar to lexically constrained translation, the model may omit some free token nonterminals (i.e., $\mathrm{Y}_n$) when generating the derivation $\mathbf{f}$, of which the ratios are 0.4\%, 0.6\%, 0.1\%, 0.9\% in English-French, English-Russian, English-Chinese, English-German, respectively. We use empty strings for the omitted nonterminals when reconstructing the output sentence.

\paragraph{Translation Performance}
Table~\ref{tab:structure-main} shows the results of all the involved methods. Our approach can improve the BLEU score over the three baselines, and the structure correctness is 100\%. Although Split-Inject can also guarantee the correctness of the output, its BLEU score is much lower, which is potentially caused by the reason that some fragments are translated without essential context. The structure match accuracy with respect to the given reference is not necessarily 100\%, since the order of markup tags can be diverse due to the variety of natural language. See Table~\ref{tab:structure-case-appendix} in Appendix for some translation examples.

\section{Conclusion} In this work, we propose a template-based framework for constrained translation and apply the framework to two specific tasks, which are lexically and structurally constrained translation.
Our motivation is to decompose the generation of the whole sequence into the arrangement of constraints and the generation of free tokens, which can be learned through a sequence-to-sequence framework.
Experiments demonstrate that the proposed method can achieve high translation quality and match accuracy simultaneously and our inference speed is comparable with unconstrained NMT baselines.

\section*{Limitations}

A limitation of this work is that our method can not cope with one-to-many constraints (e.g., $\langle$bank, \begin{CJK}{UTF8}{gbsn}河岸\end{CJK}$|$\begin{CJK}{UTF8}{gbsn}银行\end{CJK}$\rangle$). Moreover, we only validate the proposed template-based framework in machine translation tasks. However, constrained sequence generation is vital in many other NLP tasks, such as table-to-text generation~\cite{Parikh:2020:ToTTo}, text summarization~\cite{Liu:2018:Generative}, and text generation~\cite{Dathathri:2020:Plug}. In the future, we will apply the proposed method to more constrained sequence generation tasks.

\section*{Acknowledgments}

This work was supported by the National Natural Science Foundation of China (No. 61925601, No. 62006138), the National Social Science Fund of China (No. 62236011), Beijing Academy of Artificial Intelligence (BAAI), a grant from the Guoqiang Institute, Tsinghua University, and the Tencent AI Lab Rhino-Bird Focused Research Program (No. JR202031). We thank all the reviewers for their valuable and insightful comments.

\bibliography{anthology,custom}
\bibliographystyle{acl_natbib}

\clearpage
\appendix

\section{Supplementary Material for Lexically Constrained Translation}
\label{sec:lexical-appendix}

\subsection{More Details on Data}
For the lexically constrained translation task, Chinese sentences are segmented by {\tt Jieba}\footnote{\url{https://github.com/fxshy/jieba}}, while English and German sentences are tokenized using Moses~\cite{Koehn:2007:MosesOS}. The tokenized sentences are then processed by BPE~\cite{Sennrich:2016:BPE} with 32K merge operations for both the two language pairs. We detokenize the model outputs before calculating the sacreBLEU.

\subsection{More Details on Model}
We adopt Transformer~\cite{Vaswani:2017:Transformer} as our NMT model. For English-Chinese, we use the base model, whose depth is 6, and the width is 512. For English-German, we use the big model, whose depth is 6, and the width is 1024. The base and big models are optimized using the corresponding learning schedules introduced in~\citet{Vaswani:2017:Transformer}. We train base models for 200K iterations using 4 NVIDIA V100 GPUs and train big models for 300K iterations using 8 NVIDIA V100 GPUs. Each mini-batch contains approximately 32K tokens in total. All the models are optimized using Adam~\cite{Kingma:2015:CoRR}, with $\beta_{1}=0.9$, $\beta_{2}=0.98$ and $\epsilon=10^{-9}$. In all experiments, both the dropout rate and the label smoothing penalty are set to 0.1. The beam size is set to 4.

\subsection{Effect of Alignment Model}

In this work, we use an alignment model to produce word alignments for the training set, which is then used for phrase table extraction.
By default, we use all the parallel data in the training set to train the alignment model, using the {\tt fast-align} toolkit.
To better understand the effect of the alignment model, we replace the default alignment model with a weaker one that is trained using only 0.1M sentence pairs.
Table~\ref{tab:alignment} shows the result, from which we find that using the weaker word alignment can negatively affect the BLEU score. However, the exact match accuracy is still 100\%, and changes in the other two metrics are modest.

\begin{table}[t]
  \centering
  \small
  \begin{tabular}{r r r r r}
    \toprule
    \multirow{2}{*}{\bf \# Sent.} & \multirow{2}{*}{\bf BLEU} & \bf Exact & \bf Window & \multirow{2}{*}{\bf 1-TERm} \\
    & & \bf Match & \bf Overlap & \\
    \midrule
    0.1M & 37.5 & 100.0 & 32.7 & 37.5 \\
    20.6M & 38.2 & 100.0 & 32.9 & 37.6 \\
    \bottomrule
  \end{tabular}
  \caption{Effect of the alignment model on the English-Chinese validation set. ``\# Sent.'' means the number of sentence pairs used to train the alignment model.
  }
  \label{tab:alignment}
\end{table}

\subsection{Domain Robustness}

\begin{table}[ht]
  \centering
  \small
  \begin{tabular}{l r r r r}
    \toprule
    \multirow{2}{*}{\bf Method} & \multirow{2}{*}{\bf BLEU} & \bf Exa. & \bf Win. & \multirow{2}{*}{\bf 1 - T.m} \\
    & & \bf Mat. & \bf Ove. & \\
    \midrule
    Vanilla & 37.7 & 58.1 & 19.4 & 37.9 \\
    \cmidrule(lr){1-1} \cmidrule(lr){2-5}
    Placeholder & 38.5 & 98.9 & 24.4 & 38.8 \\
    VDBA & 38.0 & \bf 100.0 & 24.3 & 39.1 \\
    Replace & 38.4 & 87.3 & \underline{24.5} & 39.7 \\
    CDAlign & 38.6 & 89.3 & 24.0 & \underline{40.5} \\
    TextInfill & \underline{38.7} & 97.0 & 23.2 & 38.4 \\
    \cmidrule(lr){1-1} \cmidrule(lr){2-5}
    Ours & \bf 39.6 & \bf 100.0 & \bf 26.3 & \bf 41.3 \\
    \bottomrule
  \end{tabular}
  \caption{Results on the English-Chinese test set of the WMT21 terminology translation.}
  \label{tab:domain}
\end{table}

Domain robustness is about the generalization of machine learning models to unseen test domains~\cite{Muller:2020:DomainRobust}.
In our experiments, all the involved models are trained in the news domain.
We evaluate the domain robustness of these methods on the WMT21 terminology translation task~\cite{Alam:2021:WMT21Term}\footnote{\url{https://www.statmt.org/wmt21/terminology-task.html}}
, which lies in the COVID-19 domain. Since this task does not support English-German translation, we only conduct this experiment on English-Chinese.
In this test set, the maximum number of constraints is 12. We thus modify the phrase extraction script to increase the maximum number of constraints from 3 to 12, and then re-train both the baselines and our models. Note that we only change the number of constraints, while the training domain is still news. Since the open-sourced implementation of AttnVector~\cite{Wang:2022:Lexical}\footnote{\url{https://github.com/shuo-git/VecConstNMT}} does not support more than 3 constraints, we omit this baseline in this experiment. The test set of the WMT21 terminology translation task also contains some constraints that consist of more than one target term (i.e., one-to-many constraints). We only select the one that appear in the reference as our constraint. We leave it to future work to extend the current framework for one-to-many constraints.

Table~\ref{tab:domain} provides the results on the COVID-19 domain, where our approach performs best across all the four evaluation metrics. VDBA~\cite{Hu:2019:Lexical} and our method can both maintain the exact match accuracy, while the other three baselines achieve much lower exact match accuracy due to the domain shift. 
However, the BLEU score of VDBA is lower than other constrained translation approaches, while our method can also achieve the best BLEU score.
The exact match accuracy of TextInfill~\cite{Xiao:2022:Lexical} is lower than 100\% because sometimes the model can not generate all the slots within the length limitation. The results indicate that our approach can better cope with constraints coming from unseen domains.

\subsection{X-English Translation}

We also conduct experiments on X-English translation directions (i.e., Chinese-English and German-English). Due to the limitation of computational resources, we only train the two most recent baselines: AttnVector~\cite{Wang:2022:Lexical} and TextInfill~\cite{Xiao:2022:Lexical}. Moreover, AttnVector and TextInfill achieve the best BLEU score and exact match accuracy, excluding our approach, respectively. As shown in Table~\ref{tab:lexical-x-English}, we find that our approach performs well in both Chinese-English and German-English, achieving 100\% exact match accuracy and a better BLEU score.

\begin{table}
  \centering
  \small
  \begin{tabular}{l r r r r}
    \toprule
    \multirow{2}{*}{\bf Method} & \multirow{2}{*}{\bf BLEU} & \bf Exa. & \bf Win. & \multirow{2}{*}{\bf 1 - T.m} \\
    & & \bf Mat. & \bf Ove. & \\
    \midrule
    \cmidrule(lr){1-5}
    \multicolumn{5}{c}{\em Chinese-English} \\
    \cmidrule(lr){1-5}
    Vanilla & 23.3 & 17.6 & 10.4 & 36.6 \\
    \cmidrule(lr){1-1} \cmidrule(lr){2-5}
    AttnVector & \underline{25.9} & 95.5 & \underline{35.5} & \underline{42.1} \\
    TextInfill & 25.0 & \bf 100.0 & 33.3 & 39.0 \\
    \cmidrule(lr){1-1} \cmidrule(lr){2-5}
    Ours & \bf 26.7 & \bf 100.0 & \bf 37.3 & \bf 45.1 \\
    \midrule
    \multicolumn{5}{c}{\em German-English} \\
    \cmidrule(lr){1-5}
    Vanilla & 32.4 & 9.5 & 7.3 & 45.8 \\
    \cmidrule(lr){1-1} \cmidrule(lr){2-5}
    AttnVector & \underline{37.8} & 91.4 & 36.4 & \underline{53.3} \\
    TextInfill & 37.2 & \bf 100.0 & \underline{37.1} & 51.4 \\
    \cmidrule(lr){1-1} \cmidrule(lr){2-5}
    Ours & \bf 38.8 & \bf 100.0 & \bf 39.7 & \bf 53.4 \\
    \bottomrule
  \end{tabular}
  \caption{Results of the lexically constrained translation task in Chinese-English and German-English.}
  \label{tab:lexical-x-English}
\end{table}

\subsection{Case Study} As mentioned in Section~\ref{sec:lexical-main-res}, our approach outperforms the baselines in the lexically constrained translation task. To better understand the difference between our approach and some representative baselines, we list some examples in Table~\ref{tab:lexical-case-appendix}.

\begin{table*}[ht]
  \centering
  \small
  \scalebox{0.92}{
  \begin{tabular}{l m{14.5cm}}
  \toprule
  \bf Constraints & \begin{CJK}{UTF8}{gbsn}{\normalsize
    $\langle${\color{sred} guests },{\color{sred} 来宾 }$\rangle$};
    {\normalsize$\langle${\color{sblue} culinary culture },{\color{sblue} 食品文化 }$\rangle$};
    {\normalsize$\langle${\color{sgreen} Chinese-style },{\color{sgreen} 中式 }$\rangle$}
  \end{CJK} \\
  \cmidrule(lr){1-1} \cmidrule(lr){2-2}
  \bf Source & {\small Wang Kaiwen , Chinese ambassador to Latvia , introduced to the} {\normalsize\color{sred}guests} {\small a few major styles of cooking in Chinese gourmet foods and expressed his hope that through tasting} {\normalsize\color{sgreen}Chinese-style} {\small gourmet foods more will be learned about China and Chinese} {\normalsize\color{sblue}culinary culture}{\small .}  \\
  \cmidrule(lr){1-1} \cmidrule(lr){2-2}
  \bf Reference & \begin{CJK}{UTF8}{gbsn}
    {\small中国驻拉脱维亚大使王开文向} {\normalsize\color{sred}来宾} {\small们介绍了中国美食的几大菜系, 表示希望通过品尝} {\normalsize\color{sgreen}中式} {\small美味食品更多了解中国和中国} {\normalsize\color{sblue}食品文化} {\small。}
  \end{CJK} \\
  \cmidrule(lr){1-1} \cmidrule(lr){2-2}
  {\bf AttnVector} & \begin{CJK}{UTF8}{gbsn}
    {\small中国驻拉托维亚大使王开文向} {\normalsize\color{sred}来宾} {\small介绍了中国美食食品的几种主要烹饪方式, 并表示希望通过品尝} {\normalsize\color{sgreen}中式} {\small美食, 更多地了解中国和中国的文化。}
  \end{CJK} \\
  \cmidrule(lr){1-1} \cmidrule(lr){2-2}
  \bf TextInfill & \begin{CJK}{UTF8}{gbsn}
    {\small中国驻拉脱维亚大使王开文向}
    {\normalsize\color{sred}来宾}
    {\small介绍了几种主要的中国美食}
    {\normalsize\color{sblue}食品文化} 
    {\small, 并表示希望通过品尝}
    {\normalsize\color{sgreen}中式}
    {\small美食, 能够了解更多关于中国和中国烹饪文化的知识。}
  \end{CJK} \\
  \cmidrule(lr){1-1} \cmidrule(lr){2-2}
  {\bf Ours} & \begin{CJK}{UTF8}{gbsn}
    {\small中国驻拉脱维亚大使王开文向} {\normalsize\color{sred}来宾} {\small介绍了中国美食的几种主要烹饪风格, 并表示希望通过品尝} {\normalsize\color{sgreen}中式} {\small美食, 更多地了解中国和中国的} {\normalsize\color{sblue}食品文化} {\small。}
  \end{CJK} \\
  \midrule
  \midrule
  \bf Constraints & \begin{CJK}{UTF8}{gbsn}
    {\normalsize$\langle${\color{sred} Italian engineer},{\color{sred}义大利工程师}$\rangle$};
    {\normalsize$\langle${\color{sblue} Gidzenko},{\color{sblue}吉曾柯}$\rangle$};
    {\normalsize$\langle${\color{sgreen} Shuttleworth},{\color{sgreen}夏特沃斯}$\rangle$}
  \end{CJK}\\
  \cmidrule(lr){1-1} \cmidrule(lr){2-2}
  \bf Source & 
    {\small Returning together with}
    {\normalsize\color{sgreen} Shuttleworth} 
    {\small to earth are the Russian spacecraft commander}
    {\normalsize\color{sblue}Gidzenko} 
    {\small and the} 
    {\normalsize\color{sred}Italian engineer} 
    {\small Vittori who entered space with him.}  \\
  \cmidrule(lr){1-1} \cmidrule(lr){2-2}
  \bf Reference & \begin{CJK}{UTF8}{gbsn}
    {\small 与}
    {\normalsize\color{sgreen}夏特沃斯}
    {\small 一同返回地球的, 是这次和他一起进入太空的俄罗斯太空船指挥官}
    {\normalsize\color{sblue}吉曾柯}
    {\small 与}
    {\normalsize\color{sred}义大利工程师}
    {\small 维托利。}
  \end{CJK}\\
  \cmidrule(lr){1-1} \cmidrule(lr){2-2}
  \bf AttnVector & \begin{CJK}{UTF8}{gbsn}
    {\normalsize\color{sblue}吉曾柯}
    {\small 和}
    {\normalsize\color{sred}义大利工程师}
    {\small 维托利与}
    {\normalsize\color{sgreen}夏特沃斯}
    {\small 一同返回地球, 他们一同进入太空。}
  \end{CJK} \\
  \cmidrule(lr){1-1} \cmidrule(lr){2-2}
  \bf TextInfill & \begin{CJK}{UTF8}{gbsn}
    {\small 俄罗斯太空船指挥官吉登科(Gidzenko)和}
    {\normalsize\color{sred}义大利工程师}
    {\normalsize\color{sblue}吉曾柯}
    {\small (Vittori)与}
    {\normalsize\color{sgreen}夏特沃斯}
    {\small 一起重返地球。}
  \end{CJK} \\
  \cmidrule(lr){1-1} \cmidrule(lr){2-2}
  \bf Ours & \begin{CJK}{UTF8}{gbsn}
    {\small 与}
    {\normalsize\color{sgreen}夏特沃斯}
    {\small 一起返回地球的是俄罗斯航天器指挥官}
    {\normalsize\color{sblue}吉曾柯}
    {\small 和与他一同进入太空的}
    {\normalsize\color{sred}义大利工程师}
    {\small 维托里。}
  \end{CJK}\\
  \bottomrule
  \end{tabular}}
  \caption{Examples for lexically constrained translation. For clarity, we only list the results of two representative baselines. We choose AttnVector~\cite{Wang:2022:Lexical} and TextInfill~\cite{Xiao:2022:Lexical} since they achieve the best BLEU score and the highest exact match accuracy, respectively, excluding our approach. In the first example, AttnVector omits the target constraint {\color{sblue}\begin{CJK}{UTF8}{gbsn}食品文化\end{CJK}} in its output, while both TextInfill and our approach can generate all the three constraints. In the second example, TextInfill places the constraint \begin{CJK}{UTF8}{gbsn}{\color{sblue}吉曾柯}\end{CJK} in the wrong context, while our approach outputs a better result.}
  \label{tab:lexical-case-appendix}
\end{table*}


\section{Supplementary Material for Structurally Constrained Translation}


\subsection{More Details on Model}
\label{subsec:structure-model-appendix}

All the models are trained for 40K iterations in all the four translation directions. We adopt the cosine learning rate schedule presented in~\citet{Wu:2019:LightConv}, but we set the maximum learning rate to $7\times 10^{-4}$ and the warmup step to 8K.
The period of the cosine function is set to 32K, which means that the learning rate decays into the minimum value at the end of the training.
Both the dropout rate and the label smoothing penalty are set to 0.2. Each mini-batch consists of approximately 32k tokens in total. We use Adam~\cite{Kingma:2015:CoRR} for model optimization, with $\beta_{1}=0.9$, $\beta_{2}=0.98$ and $\epsilon=10^{-9}$. We also set the weight decay coefficient to $10^{-3}$. Both the baseline models and our models are trained using the same hyperparameters.

\subsection{Case Study}

We list some translation examples in Table~\ref{tab:structure-case-appendix} to provide a detailed understanding of our work. The examples demonstrate that our approach can effectively cope with structured inputs.

\begin{table*}[ht]
  \centering
  \small
  \scalebox{0.92}{
  \begin{tabular}{l m{14.5cm}}
  \toprule
  \bf Source & {... <ph> Each dashboard can have up to <ph> 3 </ph> filters. Contact} {\normalsize\color{sred}<ph> Salesforce </ph>} { to increase the filter options limit in <ph> Salesforce Classic </ph> . A maximum of <ph> 50 </ph> filter options is possible. </ph>} \\
  \cmidrule(lr){1-1} \cmidrule(lr){2-2}
  \bf Reference & {... <ph> Chaque tableau de bord peut inclure jusqu'\'a <ph> 3 </ph> filtres. Pour augmenter les limitations des options de filtrage dans <ph> Salesforce Classic </ph> , contactez} {\normalsize\color{sred} <ph> Salesforce </ph>} { . <ph> 50 </ph> options defiltre sont possibles au maximum. </ph>} \\
  \cmidrule(lr){1-1} \cmidrule(lr){2-2}
  \bf Split-Inject & { ... <ph> Chaque tableau de bord peut avoir jusqu'\'a <ph> 3 </ph> filtres. Contact} {\normalsize\color{sred}<ph> Salesforce </ph>} { pour accro\^itre la limitation des options de filtrage <ph> Salesforce Classic </ph> . maximum d'un maximum <ph> 50 </ph> Les options de filtrage sont possibles. </ph> L} \\
  \cmidrule(lr){1-1} \cmidrule(lr){2-2}
  \bf XML & { ... <ph> Chaque tableau de bord peut avoir jusqu'\'a <ph> 3 </ph> filtres. Pour augmenter la limitation en options de filtrage dans <ph> Salesforce Classic </ph> , chaque filtre peut inclure jusqu'\'a <ph> 50 </ph> options de filtrage. </ph>} \\
  \cmidrule(lr){1-1} \cmidrule(lr){2-2}
  \bf Ours & { ... <ph> Chaque tableau de bord peut avoir jusqu'\'a <ph> 3 </ph> filtres. Contactez} {\normalsize\color{sred} <ph> Salesforce </ph>} { pour augmenter les options de limitation de filtrage dans <ph> Salesforce Classic </ph> . Un maximum de <ph> 50 </ph> options de filtrage est possible. </ph>} \\
  \midrule
  \midrule
  \bf Source & { Each <ph> Event Monitoring app </ph> user needs an} {\normalsize\color{sred}<ph> Event Monitoring Analytics Apps </ph>} { permission set license. The} {\normalsize\color{sblue}<ph> Event Monitoring Analytics Apps </ph>} { permission set license enables the following permissions.} \\
  \cmidrule(lr){1-1} \cmidrule(lr){2-2}
  \bf Reference & { Chaque utilisateur de l' <ph> application Event Monitoring </ph> doit disposer d'une licence d'ensemble d'autorisations} {\normalsize\color{sred}<ph> Event Monitoring Analytics Apps </ph>} { . La licence d'ensemble d'autorisations} {\normalsize\color{sblue}<ph> Event Monitoring Analytics Apps </ph>} { accorde les autorisations ci-dessous.} \\
  \cmidrule(lr){1-1} \cmidrule(lr){2-2}
  \bf Split-Inject & { Chaque <ph> Application Event Monitoring </ph> utilisateur doit avoir un utilisateur} {\normalsize\color{sred} <ph> Applications Event Monitoring Analytics </ph>} { Licence d'ensemble d'autorisations.} {\normalsize\color{sblue} <ph> Applications Event Monitoring Analytics </ph>} { La licence d'ensemble d'autorisations active les autorisations ci-dessous.}\\
  \cmidrule(lr){1-1} \cmidrule(lr){2-2}
  \bf XML & { Chaque utilisateur de l' <ph> application Event Monitoring </ph> doit disposer d'une licence d'ensemble d'autorisations} {\normalsize\color{sred} <ph> Event Monitoring Analytics Apps </ph>} { . La licence d'ensemble d'autorisations} {\normalsize\color{sblue} <ph> Event Monitoring Analytics Apps </ph>} { active les autorisations ci-dessous.} \\
  \cmidrule(lr){1-1} \cmidrule(lr){2-2}
  \bf Ours & { Chaque utilisateur de l' <ph> application Event Monitoring </ph> doit disposer d'une licence d'ensemble d'autorisations} {\normalsize\color{sred} <ph> Event Monitoring Analytics Apps </ph>} { . La licence d'ensemble d'autorisations} {\normalsize \color{sblue} <ph> Event Monitoring Analytics Apps </ph>} { active les autorisations suivantes.} \\
  \bottomrule
  \end{tabular}}
  \caption{Examples for structurally constrained translation. We only highlight some text fragments wrapped by markup tags to show the difference between the involved methods. In the first example, XML~\cite{Hashimoto:2019:Structure} omits the fragment {\color{sred} <ph> Salesforce </ph>}, while Split-Inject and our method recall all the markup tags of the source sentence. In the second example, the colored contents are mistranslated by Split-Inject, which is potentially caused by the lack of context when translating these fragments.}
  \label{tab:structure-case-appendix}
\end{table*}

\end{document}